# Torque Ripple Minimization in a Switched Reluctance Drive by Neuro-fuzzy Compensation

Luís O. A. P. Henriques, *Student Member, IEEE,* Luís G. B. Rolim, Walter I. Suemitsu, *Member, IEEE,* Paulo J. C. Branco, *Member, IEEE,* Joaquim A. Dente

*Abstract*-- **Simple power electronic drive circuit and fault tolerance of converter are specific advantages of SRM drives, but excessive torque ripple has limited its use to special applications. It is well known that controlling the current shape adequately can minimize the torque ripple. This paper presents a new method for shaping the motor currents to minimize the torque ripple, using a neuro-fuzzy compensator. In the proposed method, a compensating signal is added to the output of a PI controller, in a current-regulated speed control loop. Numerical results are presented in this paper, with an analysis of the effects of changing the form of the membership function of the neuro-fuzzy compensator.**

## I. INTRODUCTION

As SR machine presents strong nonlinear characteristics, fuzzy logic and neural networks methods are well suited for its control, and so many authors have proposed the dynamic control of SR drives using these artificial intelligence based methods [2,4].

The use of fuzzy logic control has been implemented with success by the authors in [1], and has shown to be effective for the SR speed control in applications where some degree of torque ripple is tolerated, as is the case in many industrial applications. Nevertheless, in servo control applications or when smooth control is required at low speeds, the elimination of the torque ripple becomes the main issue for an acceptable control strategy. In this case, the fuzzy logic controller is not enough because torque ripple changes with the SR motor speed and load. In this context, it is advantageous to include some learning mechanism to the SR control to adapt itself to new dynamic conditions. This paper presents thus a new methodology to control a SR drive that consists on the use of a PI speed controller with the supervision of a neuro-fuzzy block responsible of torque ripple reduction.

## II. TORQUE PULSATION

With a PI-like control alone, it is not possible to obtain a ripple-free output speed at any speed range, because it would also require a ripple-free output torque for this purpose. A constant current reference can produce an oscillating torque as shown in Fig. 2.

At lower speeds, it is more convenient to compensate for the torque pulsations through phase current waveshaping. In this case, the current reference signal should vary as a function of position, speed and load torque, in order to produce the desired ripple compensation. In fact, the optimum compensating signal will be a highly non-linear function of position, speed and load. Some works [3,4] have been published, which use many different strategies to produce a compensating signal. In this work, a novel SR ripple compensation method is proposed based on [5,6], which uses a self-tuning neuro-fuzzy compensator. The proposed compensation scheme is described in the next section.

## III. PROPOSED METHOD

Fig. 1 presents a simplified block diagram of the SR-drive speed control system, showing the proposed neuro-fuzzy compensating scheme. The output signal produced by the compensator, $\Delta I_{comp}$ is added to the PI controller's output signal, $I_{ref}$, which should be ideally constant in steady state but producing significant ripple, as shown in Fig. 2. The resulting current signal after the addition, $I_{comp}$, is used as a compensated reference signal for the current-controlled SR drive converter. The compensating signal should then be adjusted in order to produce a ripple-free output torque.

The compensating signal is adjusted iteratively, through a neuro-fuzzy learning algorithm, where the training error information is derived from some internal variable of the SR drive system. In the simulation tests, the torque ripple itself has been used as the training error information. However, this approach would not be very practical for on-line implementation in a real system, since the dynamic torque is a variable that is difficult to measure. For continuous on-line training, other variables could be more appropriate, such as

Manuscript received December 14, 1999. This work was supported in part by CAPES/Brazil, FAPERJ/Brazil and ICCTI/Portugal.

L. Henriques, W. Suemitsu and L. Rolim are with Universidade Federal do Rio de Janeiro, COPPE-Elétrica Bloco H Ilha do Fundão Caixa Postal 68504 - CEP 21945-970 RJ Brasil (telephone: 55-21-260-5010, e-mail: walter@dee.ufrj.br)

P. Branco and J. Dente are with Mechatronics Laboratory, Electrical and Computers Engineering Department, Instituto Superior Técnico, Lisboa, 1049-01 Portugal (telephone: 351-21-8417432, e-mail: pbranco@alfa.ist.utl.pt)

acceleration or speed ripple. However, the torque could still be used directly in an off-line training system, e.g. for converter programming on a test rig at the factory.

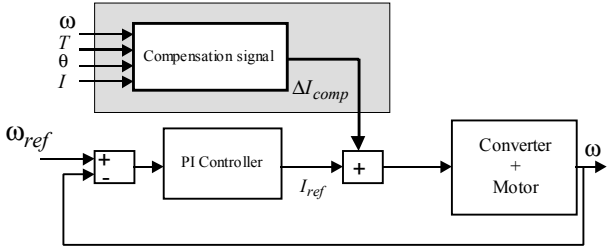

Fig. 1. Diagram of proposed SR torque ripple compensation scheme.

## IV. COMPENSATION PROCESS

The neuro-fuzzy compensator is a Sugeno-type fuzzy logic system with five fixed membership functions for each input. The types of membership functions used in this work are triangular, bell shape, and two models of gaussian shape. The rotor angular position $\theta$ and the PI controller's output signal $I_{ref}$ are used as inputs to the compensator, by means of a relation as $\Delta I_{comp} = f(\theta, I_{ref})$.

The training procedure consists on adjusting the rule consequents by a hybrid-training algorithm, which combines back-propagation and least squares minimization. At each learning iteration, the dc component is removed from the compensating signal, so that the ripple compensator does not try to change the mean value of the output torque. As a result, when the control system operates in steady-state, after the training, the PI controller will really produce a constant output signal, while the neuro-fuzzy compensator will produce a zero-mean-value compensating current reference, the $\Delta I_{comp}$ signal.

Training data are obtained from simulations of steady-state operation of the complete SR drive system. At each learning iteration, the dc component is removed from the torque signal, so that just the ripple remains. This torque ripple data is then tabulated against the mean value of the PI output reference current, and against the rotor angular position. This data set is then passed to the training algorithm, so that the torque ripple is interpreted as error information for each current-angle pair. The output of the neuro-fuzzy compensator is then readjusted to reduce the error (which is in fact the torque ripple), being this process repeated until some minimum torque ripple limit is reached.

## V. RESULTS

### A. Without Compensation

The SR motor is first controlled using only the PI regulator without compensation and full-load torque (4 Nm) at 500 rpm. Fig. 2 shows the torque signal and Fig. 3 shows its harmonic spectrum. With a 6/4 SRM, the converter produces 12 current pulses per rotor turn. So, the torque pulsations occur at a frequency 12 times higher than the frequency of rotation. For this reason, the harmonic spectrum shown in Fig. 3 exhibits non-zero components only for orders multiple of 12. The magnitudes of the harmonics are expressed as percentage of the mean value. It should be noticed that the first non-zero harmonic (12th) exhibits a quite high magnitude.

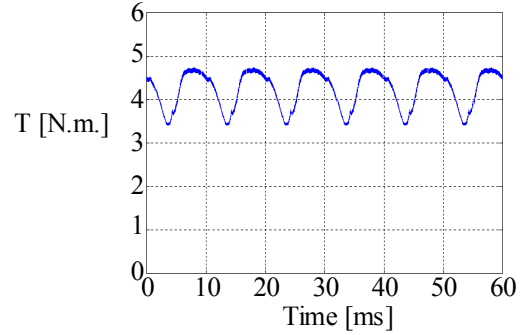

Fig. 2. Torque for non-compensated operation (500 rpm).

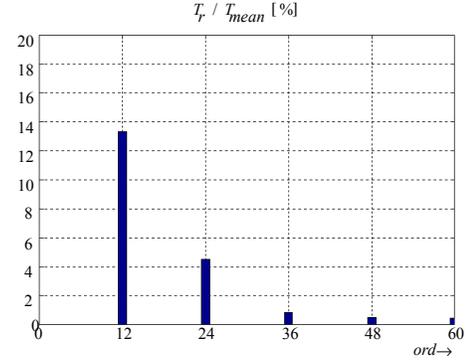

Fig. 3. Harmonics of non-compensated torque signal (500 rpm)

After ten training iterations, Fig. 4 and Fig. 5 show the output torque waveform and its harmonic content for a compensated current reference. It can be seen that the total harmonic content is very low, and the 12th harmonic is lower than 0.5% of the mean torque. After 10 training iterations, the compensated current reference produces phase current pulses like those were shown in Fig. 6. As expected, the current values are higher at the beginning and at the end of the current pulse. This pulse shape is consistent with the torque characteristics of the SR motor, which produces less torque at the beginning of pole overlapping and just before the aligned position.

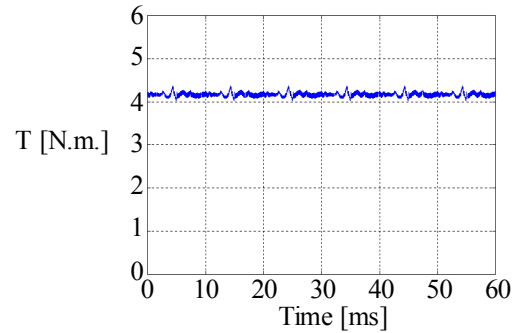

Fig. 4. Compensated torque after 10 iterations.

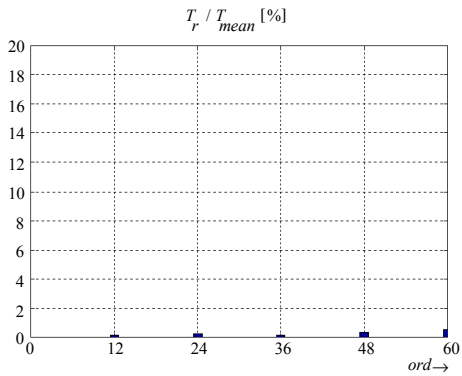

Fig. 5. Harmonic content in torque signal.

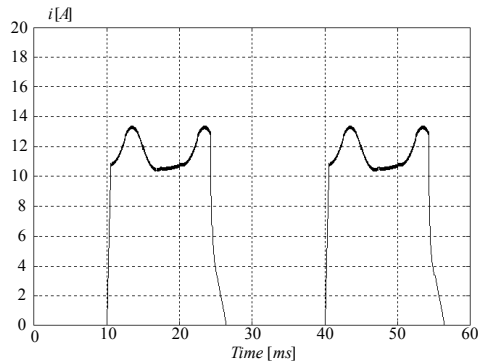

Fig. 6. Current pulses after 10 iterations.

### B. Compensation Sensitivity

Fuzzy systems have their performance significantly affected by the shape used for its membership functions. Previous results used triangular functions. The compensator performance was tested for three other membership functions: bell, and two gaussian shapes named open and normal gaussian. All results were obtained again with full-load torque, 500 rpm, and five fuzzy sets.

The results showed that using a bell shape function, the neuro-fuzzy compensator achieved its best performance. For comparison, in Fig. 7(a), we show a zoom of the harmonic content obtained for triangular functions (Fig. 5). Fig. 7(b) shows the harmonic content using the bell functions. Fig. 7(c) shows the harmonic content using the gaussian functions. We can verify that all $12^{th}$- harmonic components were decreased and not only the fundamental one as occurred with the other functions.

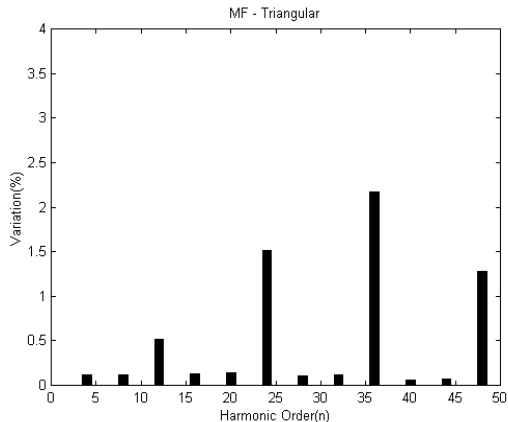

(a)

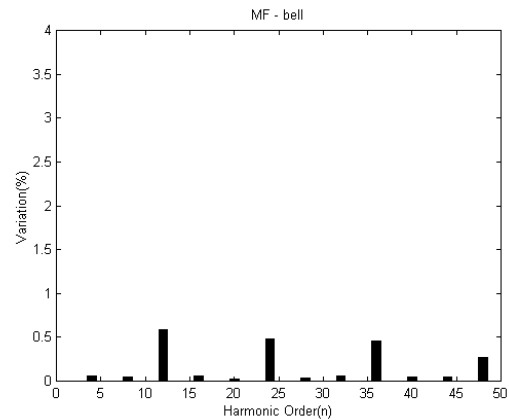

(b)

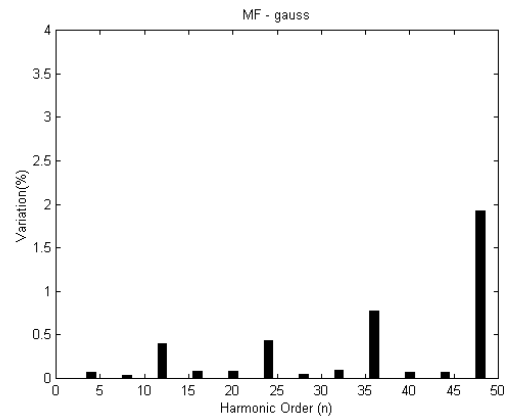

(c)

Fig. 7. Harmonics of compensated torque (500 rpm). (a) Triangular, (b) bell, (c) gaussian.

## VI. CONCLUSIONS

A neuro-fuzzy compensating mechanism to ripple reduction in SR motors was investigated. Results showed the potentialities of incorporating a compensating signal in the current waveform to minimize the torque ripple. The effect of changing the form of the membership function was also investigated revealing that a bell shape function produces better ripple reduction in all harmonic content. Next steps are the use of this concept in an experimental drive and incorporate another signal to be trained.